# Use of Deep Learning in Modern Recommendation System: A Summary of Recent Works

Ayush Singhal
Contata Solutions LLC,
Minneapolis, Minnesota, USA

Pradeep Sinha
Contata Solutions LLC,
Minneapolis, Minnesota, USA

Rakesh Pant
Contata Solutions LLC,
Minneapolis, Minnesota, USA

## ABSTRACT
With the exponential increase in the amount of digital information over the internet, online shops, online music, video and image libraries, search engines and recommendation system have become the most convenient ways to find relevant information within a short time. In the recent times, deep learning's advances have gained significant attention in the field of speech recognition, image processing and natural language processing. Meanwhile, several recent studies have shown the utility of deep learning in the area of recommendation systems and information retrieval as well. In this short review, we cover the recent advances made in the field of recommendation using various variants of deep learning technology. We organize the review in three parts: Collaborative system, Content based system and Hybrid system. The review also discusses the contribution of deep learning integrated recommendation systems into several application domains. The review concludes by discussion of the impact of deep learning in recommendation system in various domain and whether deep learning has shown any significant improvement over the conventional systems for recommendation. Finally, we also provide future directions of research which are possible based on the current state of use of deep learning in recommendation systems.

## 1. INTRODUCTION
Our day to day needs ranging from shopping items, books, news articles, songs, movies, research documents and other basic things have flooded several data-ware houses and databases both in volume and variety [1-2]. To this end, intelligent recommendation systems and powerful search engines offer users a very helpful hand. The popularity and usefulness of such systems owes to their capability to manifest convenient information from a practically infinite storehouse[3]. Thus recommendation systems such as Amazon, Netflix and similar others take initiative to know user's interest and inform users about the items of their interest. Although these systems differ from each other according to the application they are used for, the core mechanism of finding items of user's interest is that of user's interest to item matching[4].

In general, recommendations can be generated based on user preferences, item features, user-item transactions, and other environmental factors such as time, season, location. In recommendation literature these are categorized into three primary categories: collaborative filtering (using only the user-item interaction information for recommendation), content based (using user preferences, item preferences or both) and hybrid recommendation models (using both interaction information as well as user and item metadata)[5]. Models under each of these categories have their own limitations such as data sparsity, cold start for users and items[6].

Given the recent advances in the field of deep learning in various application domains such as computer vision and speech recognition, deep learning has been extended to the area of information retrieval and recommendation systems also[7]. The general opinion about the impact of integrating deep learning into recommendation system is that of significant improvement over the conventional models. In this reviews, we conduct a systematic summarization of various works pertaining to integration of deep learning into recommendation systems to provide substantial basis for reader to understand the impact and directions of future improvement of recommendation systems using deep learning.

## 2. APPROACH
In this section we describe the approach we used to collect, select and summarize research articles for this review. We used the Google scholar search engine to fetch research articles pool to select relevant papers for our review. We used the following keywords to extract articles: "recommender system deep learning", "collaborative filtering deep learning", "recurrent neural network recommender systems". We also set the time filter to "Since 2013" so that we only find articles within the last 5 years. Google Scholar fetched several articles for each query but we performed a manual selection by scanning paper titles to understand if they were actually about deep learning in recommendation systems. The manual selection left us with 33 articles. Each article was then reviewed for the deep learning approach used to enhance the recommender model and also to understand the various datasets that were used for validation purposes.

## 3. REVIEW AND DISCUSSION
Below in Fig.1, we summarize the paper counts in each of the three categorized of recommender models. We find that majority of the recent publications have leveraged deep learning to enhance collaborative filtering capabilities. In the next section, we discuss the specific deep learning techniques used for it and a brief discussion of how deep learning compliments the conventional recommender models.

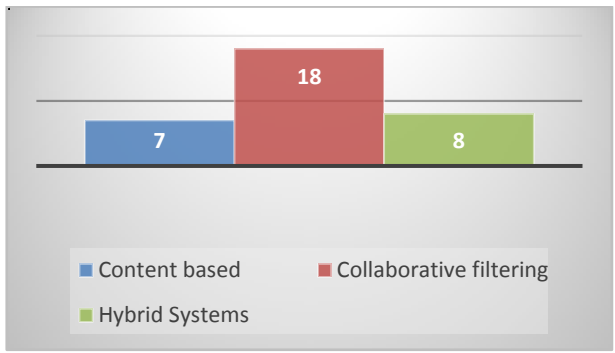

**Fig.1: Distribution of publication counts in three recommender system categories.**

In Table 1, we summarize all the publication into various application domain categories. The categorization of the articles into each domain is determined by the validation datasets used in the experiments. Typically, a recommendation system has a wider application use case but for the sake of categorization based on how the research was conducted, we have assumed the validation dataset used as the indicator of application domain. As shown in the table, most of the recent recommendation systems have been tested on datasets from the Entertainment domain. This can be attributed to the availability of public datasets from Amazon, Netflix and MovieLens. Following this, the next domain of interest for deep learning integrated recommendation systems is e-commerce. Several product review datasets from Yelp, Amazon, Slashdot etc. have helped researchers leverage such datasets for validation. The third category corresponds to the Articles such as news, quotes, and scholarly research articles. Under the social category, we place research conducted on datasets such as Epinions, social tagging etc.

**Content based system**
In this section we summarize some of the main deep learning contributions to various research works on content based recommender systems.

[8] use deep convolution neural networks to generate latent factors for songs from their audio when no usage data is available. The approach outperforms simple baseline such as linear regression, multi-layer perceptron trained on bag-of-words representation of audio signals using the Million song dataset (consisting of Last.fm dataset and Echo Nest Taste Profile Subset).

[9] proposed a combined model using deep belief networks and probabilistic graphical models to simultaneously learn features from audio content and make personalized recommendations. The model was compared with only content based models as well as hybrid model without deep learning using the Echo Nest Taste Profile Subset (a music recommendation dataset). The model outperforms both the content based baselines as well as collaborative filtering based baselines on the validation dataset.

[10][11] propose deep learning approaches for enhancing content based recommendations in the domain of quote recommendations in Writings and Dialogues. Tan et al uses LSTM to model to represent the distributed meaning of the contexts and the quotes. Lee et al combine recurrent neural network and convolution neural network to learn semantic representation of quote in the dialogue thread. Wikiquotes and Oxford Concise Dictionary of Proverbs are taken as source of quotes for tweet dialogue thread.

[12] utilize GRU based recurrent neural networks to convert item text into latent features to improve the collaborative filtering performance specially for cold start purposes. The model is tested on citation recommendation system using two real world datasets (dense and sparse versions) from CiteULike and compared with a modified version of collaborative topic modelling model. On both datasets the proposed model gave statistically significant improvement in performance.

[13] use review information to jointly learn item properties and user behaviors using deep neural networks called as Deep Cooperative Neural Networks. The model also uses a shared layer to couple the item features with user behaviors. The model is compared with 5 based lines namely, Matrix factorization, Probabilistic matrix factorization, LDA, Collaborative topic regression, Hidden Factor as Topic and collaborative deep learning using 3 real world datasets: yelp reviews, Amazon review, Beer reviews. The model outperformed all baselines on all the benchmarking datasets.

[14] have developed a news article recommendation model using dynamic attention deep model to solve the problem of non-explicit selection criteria for editors selecting news articles for the selection pool for end users. This stage of news recommendation is a step before final news recommendation for end users. In this stage the editors select a subset of news articles from a dynamically changing pool of articles getting filled from various news feed sources. There are no hard criteria for selection rules for editor selecting or rejecting certain articles. This research work uses deep learning to learn the criteria of editor's dynamic style of article selection. Such a problem is not directly solvable by conventional bag of words approaches. So the deep learning attention models are used to generate complex features to represent the article style and then classify whether the article is liked by the editor or not.

**Collaborative filtering based systems**
In this section, we summarize the main contributions of deep learning towards collaborative filtering approaches. Most of the approaches have attempted to replace matrix factorization with some version of deep neural networks.

| **Entertainment (music/ movie/ videos)** | **e-commerce** | **Articles (News/ quotes/ research articles)** | **Social** | **MISC** |
|---|---|---|---|---|
| 1. [8];<br>2. [9];<br>3. [15];<br>4. [16];<br>5. [17];<br>6. [18];<br>7. [19];<br>8. [20];<br>9. [21];<br>10. [22];<br>11. [23];<br>12. [24];<br>13. [25];<br>14. [26];<br>15. [27];<br>16. [28];<br>17. [29] | 1. [13];<br>2. [16];<br>3. [30];<br>4. [18];<br>5. [20];<br>6. [31];<br>7. [32];<br>8. [22];<br>9. [33];<br>10. [26] | 1. [10];<br>2. [11];<br>3. [12];<br>4. [14];<br>5. [15];<br>6. [34] | 1. [35];<br>2. [36];<br>3. [24];<br>4. [25];<br>5. [37] | 1. [38];<br>2. [39];<br>3. [40];<br>4. [28] |

[15] address the sparsity problem of CF approaches as well as the sparsity of auxiliary information for collaborative topic regression based approaches. The proposed model used generalized Bayesian Stacked Denoising Autoencoders. The model shows superior performance to Matrix factorization and Collaborative topic regression approaches on 2 CiteULike datasets and 1 Netflix dataset.

[16] is first of the work that provide a framework to incorporate deep learning features into CF models such as matrix factorization. The model is compared with several collaborative filtering approaches that uses matrix factorization and show some performance improvement on 4 real world datasets: MovieLens-100k, MovieLeans-1M, Book-Crossing, Advertising dataset.

[30] have developed a probabilistic rating auto-encoder to perform unsupervised feature learning and generate user profiles from a user-item rating data for enhancing collaborative filtering approaches. The results of adding deep learning to conventional collaborative filtering approaches such as matrix factorization shows statistically significant improvement in rating prediction using the yelp.com dataset. It is used in the e-commerce domain.

[17] show that Collaborative filtering can be converted to a sequence prediction problem and therefore recurrent neural network are very useful. In particular, the LSTMs are applied for CF problems and experimentally compared with k-nearest neighbor and matrix factorization approaches on two movie recommendation datasets: MovieLens and Netflix. The performance comparison shows the effectiveness of using deep learning models as compared to the other state of art models for collaborative filtering.

For session based recommendation there are several works which have used recurrent neural networks for improving recommendation. [18] first use only RNN to recommendation for short session-based data instead of long history based data. They show that traditional matrix factorization approaches are not suited for session based recommendations. Experimental evaluation on e-commerce clickstream data and YouTube-like OTT video service dataset, show a significant advantage of using RNN instead of MF approaches. [19] use item features such as image and text to further enhance RNN based session recommendations. They introduce the concept of parallel RNN that models each aspect of item namely (text, image, ID etc). Compared to simple RNN and item-kNN, the feature enriched RNN give significant improvement in performance on the YouTube like video service dataset. [20] have shown that RNN in combination with kNN are effective in improving recommendation accuracy for e-commerce application. The performance evaluation is done using public e-commerce datasets such as TMall competition and other music playlists from last.fm, artofmixing.org, 8tracks.com. [31] use the variational inference models from Bayesian statistics to improve the recurrent neural network model for session based prediction. The performance evaluation is done on e-commerce datasets and compared with several state of art approaches such as BPR-MF, GRU-based method and RNN based methods. [32] proposed a RNN model to incorporate the dwell time (the time that user spent examining a specific item) to enhance the accuracy of recommendation for session based recommendation in e-commerce datasets – Yoochoose.

[35] used deep learning to initialize user and item latent feature vectors for trust aware social recommendations and to separate the community effect in user's trusted friendships. The approach outperforms several variation of matrix factorization approaches on two real world datasets: Epinions and Flixster.

[36] have proposed a Bayesian Personalized Ranking Deep Neural Network model for friendship recommendation in social network datasets from Epinion and Slashdot. The model first extracts latent structural patterns from the input network using Convolution neural networks and then use Bayesian ranking to make recommendation. This deep learning based recommendation system outperforms the baseline approaches such as matrix factorization algorithms, Katz similarity, Adamic/Adar similarity and simple pairwise input neural network.

[38] propose a deep learning framework for long-tail web services recommendation. They use stacked denoising auto encoders to perform feature extraction for long-tail items or items which have very less content description as well as very less historical usage data. The proposed model is tested on dataset collected from the ProgrammableWeb.com and compared with several baselines such as the matrix factorization and topic models.

[21] propose a deep matrix factorization approach which uses deep neural network to map user and items into common low-dimensional space. The model uses explicit rating as well as implicit ratings. The model is compared with state of art matrix factorization approaches and tested on MovieLens movies, Amazon Music and Amazon movies datasets to show upto 7.5% improvement in NDCG metric.

[22] proposed a neural semantic personalized ranking using deep neural networks and pairwise learning to solve the cold-start problem in collaborative recommendation system. The proposed approach outperforms matrix factorization and topic regression based collaborative filtering approaches on two datasets: Netflix and CiteuLike.

[23] proposed stacked autoencoder with denoising to extract the useful low-dimension features from the original sparse user-item matrices. The approach is compared with item based CF and SVD algorithms on MovieLens dataset.

[24] developed a generalized framework using deep learning to directly model the user-item interaction matrix instead of applying deep learning only on the auxiliary data. The proposed approach completely replaces the matrix factorization based approach or matrix factorization is expressed as a special case of the proposed generic model for generating user and item latent features. The proposed generic model is compared with state of art matrix factorization approaches such as eALS and BPR and basic baselines such as ItemPop and ItemKNN on MovieLens and Pinterest datasets. The proposed approach showed consistent improvements (statistically significant) over all baselines on both datasets.

[39] leverage deep learning models esp the gated recurrent units to learn user interaction pattern to improve recommendation for personalized adaptive user interfaces. The comparative evaluation of the deep learning approach against the state of art tensor factorization and metric embedding methods shows that the deep learning approach outperforms the state of art approaches on user-interface, web-browsing and e-learning datasets.

[25] proposed a graph convolution matrix completion approach which uses graph auto-encoder framework for matrix completion. In addition to only using the interaction information of users and items, the model is generalizable to include side information of both users and items. The model's performance in tested comprehensively on 6 real world datasets: Flixster, Douban, YahooMusic, MovieLens100K, MovieLens1M, MovieLens10M and compared with other matrix completion approaches such as geometric matrix completion, alternating least square optimization method, CNN based matrix completion methods.

**Hybrid system**
In this section, we summarize the main contributions of deep learning towards hybrid recommender systems.

[40] developed two deep neural network based methods to improve personalized tag-aware recommendation. The model utilizes both the user and item profiles based on tags and convert them into common latent space using the deep learning models. The research finds that the proposed models outperformed traditional recommendation approaches, such as clustering based cosine similarity, clustering based collaborative filtering and auto-encoder based collaborative filtering approaches, with a statistically significant percentage.

[33] developed a convolution neural network based model to incorporate the meta-data information of users or items to improve the matrix factorization approach. The model is evaluated on movie lens and Amazon review datasets and compared with several state of art collaborative filtering approaches.

[26] proposed a denoising Auto-encoders based collaborative filtering approach. The model is presented as a generalized framework for all collaborative filtering approaches but more flexible for tuning. The model is compared outperform baselines such as ItemPop, ItemCF, Matrix factorization, BPR, and FISM on MovieLens, Yelp and Netflix datasets.

[37] have developed a deep -semantic matrix factorization model to improve the performance of tag-aware personalized recommendation. In their work they have integrated the techniques of deep semantic modeling, hybrid learning and matric factorization to gain improvement in performance. Experimental analysis on the real world social bookmarking dataset from Delicious bookmarking system show that the deep learning enhanced matrix factorization approach gives significant improvement in the tag recommendation task as compared with clustering based models, matrix factorization, encoder based model and deep semantic similarity based methods.

[34] propose a deep neural network with attention to tackle the problem of end user recommendation. The architecture is proposed to handle the problem of dynamic nature of news readership where the interests of the users are dynamically changing with time. The model handles both the status users and item features as well as the dynamic reading interests of the users using attention based recurrent neural networks. Extensive experiments on real world dataset from CLEF NewsREEL, this research establishes the effective of deep learning over other baselines such as Item popularity and matrix factorization.

[27] provide a deep learning based architecture for hybrid recommendation system. The approach uses doc2vec model to represent user and item profiles and predict the relevance of an item for the user using a classifier. The collaborative component of the approach uses a k-nearest neighbor approach for predicting the ratings of an item for a user. Further the results of content based and collaborative based models are combines using Feedforward neural network. The performance is tested on the MovieLens dataset.

[28] propose a deep learning framework to utilize item and user side information to support the lack of information from the sparse user-item rating matrix. For side information utilization additional Stacked Denoising Autoencoder (aSDAE) are used to convert it to latent dimensions and combined with matrix factorization latent factor matrices. The model outperforms several state of art algorithms for collaborative filtering when tested on the MovieLens dataset and Book-Crossing dataset (book rating by readers).

[29] propose a context-aware hybrid model which integrates convolution neural network into probabilistic matrix factorization with statistics of items. This approach capture

contextual information and consider Gaussian noise differently. The approach compares with non-deep learning based approaches such as matrix factorization and Collaborative topic regression as well as deep learning approaches such as collaborative deep learning using 3 real world datasets: 2 movie lens datasets and 1 Amazon instant video dataset to show slightly better performance.

## 4. CONCLUSION AND DIRECTIONS FOR FUTURE WORK

In this review, we systematically summarized various deep learning developments in the area of recommendation systems. We discussed various research works under three main categories of recommendation system: content based, collaborative filtering based and hybrid systems. We find that most deep learning efforts have been towards enhancing collaborative filtering approaches and have shown significant improvement over the state of art matrix factorization approaches. We also find that most of the deep learning development has been biased towards entertainment industry such as in movie and music recommendation. This can be largely attributed to the availability of rich datasets for validation.

Moving forward in this direction, we expect that further advancement of recommender system can happen in the following ways:

1. Creating public datasets in other application domains such as scholarly author-article datasets, online retail shopping datasets, and other datasets which contain both user-item interaction as well as rich metadata content about user and items.
2. Creating user inclusive test bed for evaluating recommender system performance improvement in near-real world settings. Currently the improvements using deep learning has been marked in the range of 5-8% in most of the above research works. But these improvements need to be tested in real world setting as well. It could be also accomplished by post-deployment analysis of the revenue or engagement generated by an industry's recommender system with deep learning integration.
3. There is a need of a meta-analysis which compares all the deep learning models over same set of benchmarking datasets.

## 5. ACKNOWLEDGMENTS